\documentclass{article}

\usepackage[T1]{fontenc}
\usepackage{hyperref}
\usepackage{microtype}

\usepackage{subfigure}
\usepackage{wrapfig}
\usepackage[final,nonatbib]{neurips_2021}
\usepackage{amsthm}
\newtheorem{theorem}{Theorem}
\newtheorem{proposition}[theorem]{Proposition}

\newtheorem{corollary}[theorem]{Corollary}

\newtheorem{definition}[theorem]{Definition}

\newtheorem*{remark}{Remark}

\makeatletter
\renewcommand{\@noticestring}{Optimal Transport and Machine Learning Workshop at NeurIPS 2021}
\makeatother

\newtheorem{innercustomprop}{Proposition}
\newenvironment{customprop}[1]
{\renewcommand\theinnercustomprop{#1}\innercustomprop}
{\endinnercustomprop}

\usepackage{AVTcommands}
\usepackage{AVTcitations}
\usepackage{AVTenvironments}

\bibliography{references}

\author{Samuel Cohen\textsuperscript{\ensuremath{1}}
\\\bfseries Brandon Amos\textsuperscript{\ensuremath{4}}
\And
Alexander Terenin\textsuperscript{\ensuremath{2,5}}
\\\bfseries Marc Peter Deisenroth\textsuperscript{\ensuremath{1}}
\And
Yannik Pitcan\textsuperscript{\ensuremath{3}}
\\\bfseries K S Sesh Kumar\textsuperscript{\ensuremath{5}}
\AND
\normalfont
\textsuperscript{\ensuremath{1}}Centre for Artificial Intelligence, University College London
\quad 
\textsuperscript{\ensuremath{2}}University of Cambridge 
\\
\textsuperscript{\ensuremath{3}}University of California, Berkeley
\quad 
\textsuperscript{\ensuremath{4}}Facebook AI Research
\quad
\textsuperscript{\ensuremath{5}}Imperial College London
}
\title{Sliced Multi-Marginal Optimal Transport}

\begin{document}

\maketitle

\begin{abstract}
Multi-marginal optimal transport enables one to compare multiple probability measures, which increasingly finds application in multi-task learning problems.
One practical limitation of multi-marginal transport is computational scalability in the number of measures, samples and dimensionality.
In this work, we propose a multi-marginal optimal transport paradigm based on random one-dimensional projections, whose (generalized) distance we term the \emph{sliced multi-marginal Wasserstein distance}.
To construct this distance, we introduce a characterization of the one-dimensional multi-marginal Kantorovich problem and use it to highlight a number of properties of the sliced multi-marginal Wasserstein distance. 
In particular, we show that (i) the sliced multi-marginal Wasserstein distance is a (generalized) metric that induces the same topology as the standard Wasserstein distance, (ii) it admits a dimension-free sample complexity, (iii) it is tightly connected with the problem of barycentric averaging under the sliced-Wasserstein metric.
We conclude by illustrating the sliced multi-marginal Wasserstein on multi-task density estimation and multi-dynamics reinforcement learning problems.
\end{abstract}

\section{Introduction}

Optimal transport is a framework for defining meaningful metrics between probability measures \cite{villani, compopt}. 
These metrics find a wide range of applications, such as generative modeling \cite{pmlr-v84-genevay18a,bunne2019}, Bayesian inference \cite{bayes}, imitation learning \cite{Dadashi2020PrimalWI}, graph matching and averaging \cite{pmlr-v97-xu19b,NIPS2019_8569}.
Multi-marginal optimal transport \cite{gangbo} studies ways of comparing more than two probability measures in a geometrically meaningful way.
Multi-marginal distances defined using this paradigm are often useful in settings where sharing geometric structure is useful, such as multi-task learning. In particular, they have been applied for training multi-modal generative adversarial networks \cite{mwgan}, clustering  \cite{generalizedmet}, and computing barycenters of measures \cite{altschuler_mm}.

Following the establishment of key theoretical results, including by \textcite{gangbo,journals/siamma/AguehC11, Pass2014MultimarginalOT}, research is shifting toward applications. This motivates a need for practical algorithms for the multi-marginal setting \cite{mw_compl}. 
Standard approaches based on linear programming and entropic regularization scale exponentially with the number of measures, and/or the dimension of the space \cite{benamou:hal-01096124, accmmot}. 
A number of recent works have therefore studied settings, where multi-marginal transport problems can be efficiently solved via low-rank structures on the underlying cost function \cite{altschuler_mm}, but exponential cost in the dimension remains \cite{altschulernpbary, altschuler_mm_np}.

In parallel, a number of works on \emph{sliced transport} \cite{bonnottee} developed techniques for scalable transport, which (i) derive a closed form for a problem in a single dimension, and (ii) extend it into higher dimensions via random linear projections (slicing) and thereby inherit the complexity of the one-dimensional problem.
This strategy has been shown effective in the classical Wasserstein \cite{bonnottee,bonneel, gensliced, distributionalsliced, maxsliced, orthsliced} and Gromov--Wasserstein \cite{sliced_gw} settings between pairs of measures, but has not yet been applied to settings with more than two measures.

In this paper, we address this gap and propose \emph{sliced multi-marginal transport}, providing a scalable analog of the multi-marginal Wasserstein distance.
To do so, we derive a closed-form expression for multi-marginal Wasserstein transport in one dimension, which lifts to a higher-dimensional analog via slicing. 
This one-dimensional closed-form expression can be computed with a complexity of $\c{O}(PN\log N)$, where $P$ is the number of measures and $N$ is the number of samples per measure. 
Sliced multi-marginal Wasserstein ($\c{SMW}$) can be estimated by Monte Carlo in $\c{O}(KPN\log N)$, where $K$ is the number of Monte Carlo samples. 

Furthermore, we study $\c{SMW}$'s theoretical properties.
We prove that (i) it is a generalized metric, whose associated topology is the topology of weak convergence, (ii) its sample complexity is dimension free, just like the sliced Wasserstein case involving two measures, and (iii) sliced multi-marginal transport is closely connected with the problem of barycentric averaging under the sliced Wasserstein metric. 
%
We also showcase applications, where we focus on multi-task learning on probability spaces, where sharing knowledge across tasks can be beneficial and sliced multi-marginal Wasserstein can be used as a regularizer between task-specific models.

\section{Background}
\label{sec:background}

Multi-marginal optimal transport \cite{gangbo} is a class of optimization problems for comparing multiple measures $\mu_1,\ldots,\mu_P \in \c{M}(\R^d)$, all supported on the metric space $(\R^d,  || \cdot||_2)$. 
The most common such problem is computing the multi-marginal Wasserstein distance, defined as
\[
\label{eq:multi_marginal_w}
\mathcal{MW}^2(\mu_1,\ldots,\mu_P)=\hspace{-4mm}\min_{\pi \in \Pi(\mu_1,\ldots, \mu_P)} \int_{(\R^d)^P}  c(\v{x}_1,\ldots,\v{x}_P) \d\pi(\v{x}_1,\ldots,\v{x}_P),
\]
where $c:\R^d \times\ldots\times \R^d \-> \R$ is a cost function and $\Pi(\mu_1,\ldots, \mu_P)$ is the set of probability measures in $\c{M}((\R^d)^P)$ with marginals $\mu_1, \ldots, \mu_P$.
We focus on the barycentric cost of \textcite{gangbo, journals/siamma/AguehC11}, given by
\[
c(\v{x}_1,\ldots,\v{x}_P) = \sum_{p=1}^P \beta_p \Big\Vert  \v{x}_p - \sum_{j=1}^P\beta_j\v{x}_j\Big\Vert ^2,
\quad\beta_1,\ldots,\beta_P \geq 0, \quad \sum_{p=1}^P \beta_p = 1
.
\]
This cost was originally motivated from an economics-inspired perspective, but is also often preferable because it leads to connections with barycentric averaging \cite{journals/siamma/AguehC11}, giving it a simple interpretation.
It also recovers the Wasserstein distance with squared $2$-Euclidean cost in the case $P=2$ (up to constants), referred to as $\c{W}$.
Algorithms for estimating \eqref{eq:multi_marginal_w} from a set of samples  scale exponentially with the number of measures $P$ and/or the dimension $d$ of the ground space  \cite{altschuler_mm, altschulernpbary, benamou:hal-01096124}.

$\c{MW}$ is useful in multi-task settings for regularizing measures $\mu_1,\ldots, \mu_P$ by adding $\c{MW}(\mu_1,\ldots, \mu_P)$ to a multi-task loss. 
It can also be used in a setting, where we aim for a model output $\mu$ to be close to a given set of measures $\nu_1,\ldots, \nu_P$, which can be done by introducing a loss of the form $\c{MW}(\mu, \nu_1,\ldots, \nu_P)$ and minimizing it with respect to $\mu$.


\begin{figure}
\centering
\includegraphics[height=3cm]{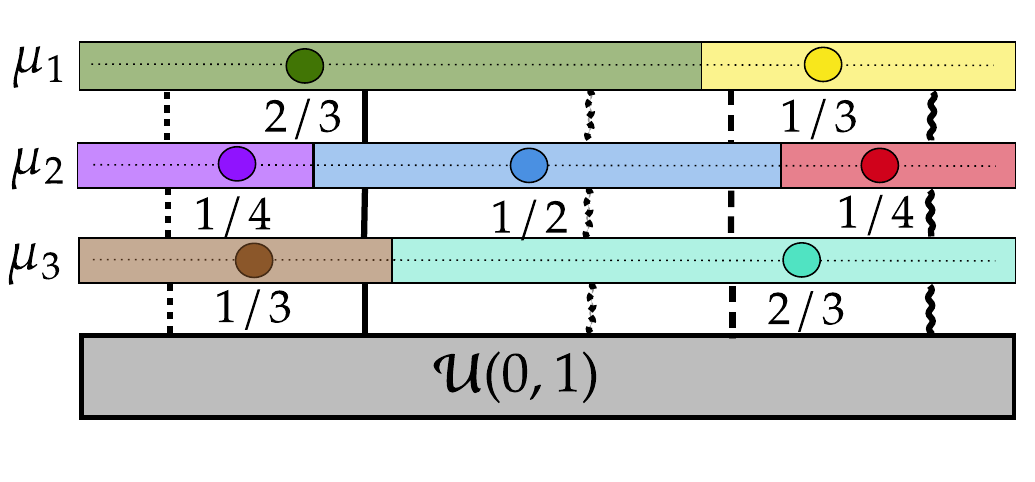}
\hspace{5mm}
\includegraphics[height=3cm]{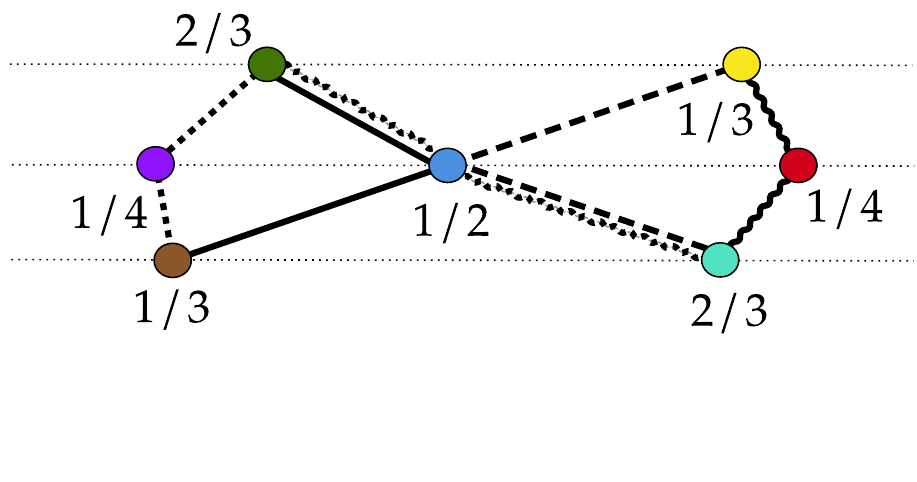}
\caption{Illustration of the optimal coupling's structure on $\R$ between discrete measures $\mu_1, \mu_2$ and $\mu_3$. Points are samples of each measures, with weights next to them.  Left: histogram of measures (horizontal); joint samples are obtained by sampling a (black) line uniformly (drawn vertically), and picking points that are associated with the bin intersected by that line. Right: Corresponding triples of points that are aligned according to the coupling are linked by a pair of lines.}
\label{fig:illustr_mmot}
\end{figure}
\textbf{Sliced transport.}  With the usual Euclidean-type cost structures, the Wasserstein distance between pairs of one-dimensional discrete measures can be computed efficiently using \emph{sorting}  with $\c{O}(N \log N)$ complexity.
More generally, we can consider the average distance between measures projected onto $\R$ along random axis, which gives \cite{bonnottee, bonneel}
\[
\c{SW}^2(\mu, \nu) = \int_{S_{d-1}}\c{W}^2\big(M^{\v{\theta}}_{\#}(\mu), M^{\v{\theta}}_{\#}(\nu)\big) \d\Theta(\v{\theta}),
\]
where $M^{\v{\theta}}(\v{x}) = \v{x}^T\v{\theta}$, $(\.)_\#$ denotes the push-forward of measures, and $\Theta$ is the uniform distribution on the unit sphere $S_{d-1}$. We sample from $M^{\v{\theta}}_{\#}(\mu)$ by sampling from $\mu$ and  projecting onto $\v{\theta}$.

A fundamental result by \textcite{bonnottee} is that $\c{SW}$ is a metric that metrizes the topology of weak convergence---the \emph{exact same} topology as $\c{W}$.
$\c{SW}$ can be estimated via Monte Carlo and preserves the computational complexity of estimating $\c{W}$ on $\R$, which is $\c{O}(N\log N)$.
Owing to the Monte Carlo nature, the sample complexity of $\c{SW}$ is dimension free \cite{bonnottee, topstatprop}, in contrast with the exponential dependency of the Wasserstein distance on dimension.
The combination of good computational and statistical properties makes $\c{SW}$ an attractive choice for  minimization problems on measure spaces, including generative modeling and imitation learning \cite{maxsliced, Dadashi2020PrimalWI}.
This immediately raises the question whether $\c{SW}$ extends to the multi-marginal case so that it preserves its key appealing properties.

\section{Sliced Multi-Marginal Transport}

To proceed toward a suitable notion of sliced multi-marginal optimal transport, we begin by developing a probabilistic analogy to understand the coupling structure that arises in one-dimensional transport when considering multiple measures.
This enables us to derive suitably-closed-form expressions from which sliced multi-marginal Wasserstein distances can be built.

\subsection{One-dimensional Multi-Marginal Transport}

In optimal transport, couplings between probability measures form one of the standard objects of study.
One way to understand the structure of a coupling is to introduce a set of random variables $y_i : \Omega\->\R$ on a probability space $(\Omega,\c{F},\P)$ whose joint distribution is the coupling of interest.
Consider the one-dimensional Wasserstein formula
\[
\c{W}^2(\mu_1,\mu_2) = \int_0^1 |C_{\mu_1}^{-1}(x) - C_{\mu_2}^{-1}(x)|^2 \d x,
\label{eq:W^2 init}
\]
where $C_{\mu_1}^{-1}, C_{\mu_2}^{-1}$ are the generalized quantile functions of $\mu_1, \mu_2$.
If we define $y_1 = C_{\mu_1}^{-1}(x)$ and $y_2 = C_{\mu_2}^{-1}(x)$, taking $\big([0,1],\c{B}(0,1),\c{U}(0,1)\big)$ as our probability space, we can write \eqref{eq:W^2 init} as
\[
\c{W}^2(\mu_1,\mu_2) &= \E_{y_1,y_2 \~ \Pi} \big[|y_1 - y_2|^2\big]
&
\Pi &= (C_{\mu_1}^{-1}, C_{\mu_2}^{-1})_\#\ \c{U}(0,1)
.
\]
This reveals that the optimal coupling admits a very specific structure: it is the pushforward measure induced by an underlying uniform random variable.
The one-dimensional Wasserstein distance is thus simply the average squared distance of a pair of random variables $y_1$ and $y_2$, where (a) we sample both $y_1$ and $y_2$ by the generalized quantile method, and (b) we \emph{share the underlying uniform random numbers} used in the sampling.
We prove that this view is general and extends to the multi-marginal case, even in the case of the more elaborate cost structure introduced in Section \ref{sec:background}.

\begin{proposition}
\label{prop:closed_form}
If $\mu_1,\ldots, \mu_P \in \c{M}(\mathbb{R})$ and $\c{U}(0,1)$ is the uniform measure, then
\[
\label{eq:mmwass_closed}
        \mathcal{MW}^2(\mu_1,\ldots,\mu_P)&=\int_{0}^1\sum_{p=1}^{P}\beta_p\Big|C_{\mu_p}^{-1}(x)-\sum_{j=1}^P\beta_jC_{\mu_j}^{-1}(x)\Big|^2\d x,
\]
and the optimal coupling solving \eqref{eq:multi_marginal_w} is of the form
\[
\label{eq:optimalplan}
        \pi^\star = (C_{\mu_1}^{-1},\ldots, C_{\mu_P}^{-1})_{\#}\ \c{U}(0,1).
\]
\end{proposition}


Proposition \ref{prop:closed_form} shows the optimal coupling is the push-forward of a uniform distribution through the generalized quantiles of each measure. Obtaining joint samples from the coupling can hence be done by sampling from the uniform distribution and mapping through each quantile function.
This extends the result by \textcite{carlier:hal-00987292} to the setting where absolute continuity is not assumed.
In the discrete case, we can simplify this further by introducing the sorting idea used in the one-dimensional Wasserstein case, to deduce the following.

\begin{corollary}
\label{prop:closedformdiscrete}
If measures $\mu_1,\ldots, \mu_P \in \c{M}(\R)$ are discrete and uniform with $N$ atoms, i.e.,  $\mu_p =\frac{1}{N} \sum_{i=1}^N\delta_{\tilde{x}_i^{(p)}}$, with $\tilde{x}_1^{(p)}\leq \ldots \leq \tilde{x}_N^{(p)}$, for $p=1,\ldots,P$, then 
\begin{align}
\label{eq:mmwasscl}
     \mathcal{MW}^2(\mu_1,\ldots,\mu_P)&=\frac{1}{N}\sum_{i,p=1}^{N,P}\beta_p\Big|\tilde{x}_i^{(p)}-\sum_{j=1}^P\beta_j\tilde{x}_i^{(j)}\Big|^2.
\end{align}

\end{corollary}

In particular, this means that the complexity of computing the multi-marginal Wasserstein in one dimension in the discrete uniform case is $\c{O}(PN\log N)$---the cost of sorting.
This establishes the necessary results in one dimension, and we generalize them to the higher-dimensional case via slicing.

\subsection{Sliced Multi-Marginal Wasserstein Distance}

To define the sliced multi-marginal Wasserstein distance, we average the expressions given in \eqref{eq:mmwass_closed} along  one-dimensional random projections, which gives
\[
\label{eq:smmwass}
        \mathcal{SMW}^2(\mu_1,\ldots,\mu_P)&= \int_{S_{d-1}} \int_{0}^1\sum_{p=1}^{P}\beta_p\Big| C_{\mu_p^{\v{\theta}}}^{-1}(x)-\sum_{j=1}^P\beta_j   C_{\mu_j^{\v{\theta}}}^{-1}(x) \Big|^2 \d x \d \Theta(\v{\theta}),
\]
where $ \mu_j^{\v{\theta}} = M_{\v{\theta}\#}(\mu_j)$ for $ j=1,\ldots,P$.
$\c{SMW}$ in \eqref{eq:smmwass} can be estimated via Monte Carlo in $O(KPN\log N)$, where $K$ is the number of Monte Carlo samples (projections). 

\paragraph{Topological properties}
We now study $\c{SMW}$'s  topological properties. We first show that  $\c{SMW}$  is the weighted mean of sliced Wasserstein distances between pairs of measures.

\begin{proposition}
Let $\mu_1,\dots,\mu_P \in \c{M}(\R^d)$. We have that
\[
\c{SMW}^2(\mu_1, \ldots, \mu_P)  = \frac{1}{2} \sum_{i,j=1}^P\beta_i \beta_j \c{SW}^2(\mu_i,\mu_j)
.
\label{eq:meansliced}
\] 
\label{prop:meansliced}
\end{proposition}
Proposition \ref{prop:meansliced} is useful in deriving statistical and topological properties of $\c{SMW}$. It is however more efficient to estimate it via our closed-form formula for multi-marginal transport -- see \eqref{eq:smmwass}. This leads to a computational complexity of $O(KPN\log N)$, whereas naively implementing \eqref{eq:meansliced} scales in $\c{O}(KP^2N\log N)$. 
Furthermore, as the sliced-Wasserstein metric is upper-bounded by the Wasserstein \cite{bonnottee}, an immediate consequence of Proposition \ref{prop:meansliced} is that 
\[
\c{SMW}^2(\mu_1, \ldots, \mu_P) \stackrel{\eqref{eq:meansliced}}{=} \frac{1}{2} \sum_{i,j=1}^P\beta_i \beta_j \c{SW}^2(\mu_i,\mu_j) \leq \frac{1}{2} \sum_{i,j=1}^P \beta_i \beta_j\c{W}^2(\mu_i,\mu_j).
\] 

A reverse inequality also follows directly (see corollary \ref{cororeverse}), which shows that $\c{SMW}$ gives rise to the topology of weak convergence---one of the key properties that made $\c{SW}$ an attractive choice in the first place. 
We now study metric properties of $\c{SMW}$.

\begin{proposition}
$\c{SMW}$ is a generalized metric.
\label{prop:metricprop}
\end{proposition}


In particular, this means that $\c{SMW}$ is (i) non-negative, (ii) zero if and only if all measures are identical, (iii) permutation-equivariant, and (iv) satisfies a generalized triangle inequality involving multiple measures. Hence, $\c{SMW}$ is well-behaved topologically-wise as it is a generalized metric inducing weak convergence. We continue by studying $\c{SMW}$'s statistical properties.

\paragraph{Statistical Properties}
In the following proposition, we assess the impact of the number of samples and random projections used to estimate $\c{SMW}$.

\begin{proposition}
\label{prop:samplecomplexity}
If $\mu_1,\ldots,\mu_P \in \c{M}(\R^d)$, and assuming $\c{W}^2$ has sample complexity $\rho(N)$ on $\R$, then,
\[ &\quad \E [\c{SMW}^2(\mu_1,\ldots,\mu_P) - \c{SMW}^2(\h{\mu}_1,\ldots,\h{\mu}_P)]^2 
\leq\frac{1}{2}\rho(N),
\]
where $\h{\mu}_p$ refers to empirical measures with $N$ samples.
\end{proposition}


Proposition \ref{prop:samples} shows that the sample complexity of $\c{SMW}$ is dimension-free---this stands in contrast to the sample complexity of the multi-marginal Wasserstein, which is exponential in the dimension.
In practice, we use Monte Carlo sampling to compute $\c{SMW}$, which introduces additional error.
To understand this error, we examine $\c{SMW}$'s projection complexity.

\begin{proposition}
\label{prop:projectioncomplexity}
Let $\mu_1,\ldots,\mu_P \in \c{M}(\R^d)$, and define $\widebar{\c{SMW}}$ the approximation obtained by uniformly picking $L$ projections on $S_{d-1}$, then
\[ 
\E\left[ \widebar{\c{SMW}}^2(\mu_1,\ldots,\mu_P) - \c{SMW}^2(\mu_1,\ldots,\mu_P)\right]^2
\leq 
L^{-1/2}\Var_{\v{\theta}}\Big[\c{MW}^2\big(\mu_1^{\v{\theta}},\ldots,\mu_P^{\v{\theta}})\Big],
\]
where  $\v{\theta}$ follows the uniform distribution on $S_{d-1}$ and $\mu_p^{\v{\theta}} = M_{\#}^{\v{\theta}}(\mu_p)$.

\end{proposition}


This shows that the quality of Monte Carlo estimates of $\c{SMW}$ is controlled by number of projections and the variance of evaluations of the base multi-marginal Wasserstein in 1D.

\paragraph{Connection to Barycenters}

We now study connections of $\c{SMW}$ to the problem of barycentric averaging, which extends the notion of a \emph{mean} to more general settings. 
Let $\c{D}:\c{M}(\R^d)\times \c{M}(\R^d) \-> \R$ be a discrepancy on the space of probability measures. Recall that the \emph{barycenter} of $P$ measures $\mu_1,\ldots,\mu_P$ is defined as
\[
\mu^\star &= \argmin_{\mu \in \c{M}(\R^d)} \c{F}(\mu),
&
\c{F}(\mu) &=\sum_{p=1}^P\c{D}(\mu_p, \mu).\label{eq:baryaveraging}
\]
Barycentric averaging is well-studied from  theoretical and computational view-points, notably under the squared Wasserstein \cite{pmlr-v32-cuturi14}, sliced Wasserstein \cite{bonneel} and Gromov--Wasserstein \cite{gwaveraging} metrics.

\begin{proposition}\label{prop:equiv}
Let $\mu_1,\ldots,\mu_P\in \c{M}(\mathbb{R}^d)$, $\sum_{p=1}^P \beta_p=1$. Furthermore, let $\hat{\beta}_p$ be augmented multi-marginal weights, so that for $m \in [0,1]$ it holds that $\hat{\beta}_p = m \beta_p$ for $p=1,\ldots,P$, $\sum_{p=1}^{P+1}\hat{\beta}_p = 1$, and $\c{D}=\c{SW}^2$. Then
\[
    \argmin_{\mu \in \c{M}(\mathbb{R}^d)} \c{SMW}^2(\mu_1,\ldots,\mu_P, \mu) = \argmin_{\mu \in \c{M}(\mathbb{R}^d)}\c{F}(\mu),
\]
where $\v{\beta}$ is the weight vector of $\c{F}$ and $\v{\hat{\beta}}$ is the weight vector of $\c{SMW}$.
\label{prop:connectionbary}
\end{proposition}


Proposition \ref{prop:connectionbary} reveals a connection between sliced multi-marginal transport and barycenters under the sliced-Wasserstein: the measure that is closest to $\mu_1, \ldots, \mu_P$ in $\c{SMW}$ is actually the barycenter of such measures under $\c{SW}$. We continue by studying smoothness of $\c{SMW}$ as a loss function.

\paragraph{Differentiability}

Sliced Wasserstein variants are desirable candidate losses for learning on probability spaces thanks to their  smoothness properties. We show $\c{SMW}$ inherits these properties.

\begin{proposition}
\label{prop:differentiability}
Let $\mu_1, \ldots, \mu_P \in \c{M}(\R^d)$ be discrete measures with $N$ atoms, which we gather into matrices $\{\m{X}^{(p)}\}_{p=1}^P$, and similarly define $\mu_{\m{X}}$ with atoms $\m{X}$. Assume $\m{X}$ has distinct points. Then $\c{SMW}^2$ is smooth with gradient
\[
\nabla_{\m{X}} \c{SMW}^2(\mu_1,\ldots,\mu_P, \mu_{\m{X}}) =\beta_{P+1}\sum_{p=1}^{P} \beta_{p}\int_{S_{d-1}}\m{X}_{\v{\theta}}-\big(\m{X}_{\v{\theta}}^{(p)} \circ \sigma_{\m{X}_{\v{\theta}}} \circ \sigma^{-1}_{\m{X}^{(p)}_{\v{\theta}}} \big) \d\Theta(\v{\theta}),
\]
where $\sigma_{\m{X}}$ is the permutation that sorts atoms of $\m{X}$.
\end{proposition}

\begin{figure}
  \begin{center}
    \includegraphics[width=0.75\hsize]{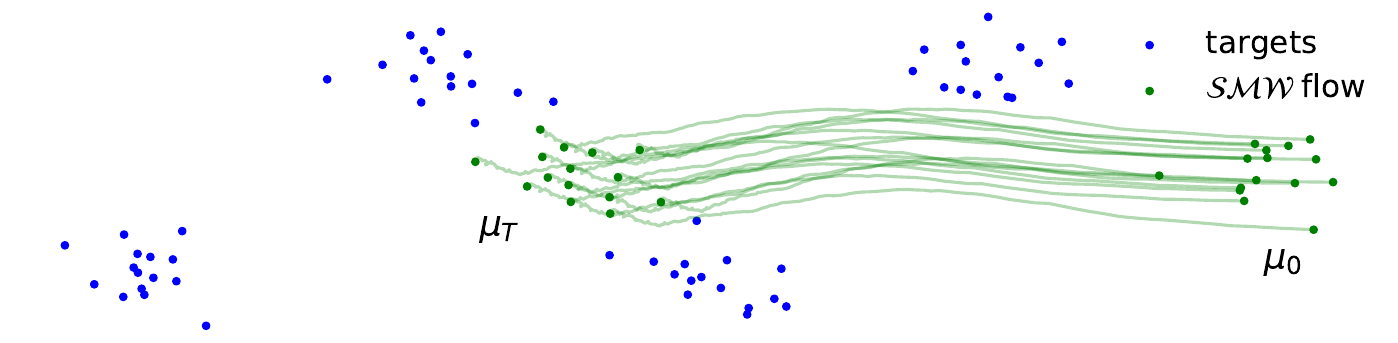}
  \end{center}
  \caption{Gradient flow $\partial \mu_t = -\nabla \c{SMW}^2(\mu_t, \nu_1,\ldots, \nu_P)$ starting from a randomly initialized Gaussian $\mu_0$. It is solved iteratively following \textcite{bonneel}.}
  \label{fig:differentiability}
\end{figure}

Proposition \ref{prop:differentiability} shows that $\c{SMW}^2$ is smooth almost everywhere, and is hence well-suited for multi-task learning, as it allows to compare multiple task-representative probability measures.
We illustrate this in Figure \ref{fig:differentiability}. Here, we consider the problem $\min_{\mu}\c{SMW}^2(\mu, \nu_1,\ldots,\nu_4)$,  amounting to estimating the sliced barycenter of $\mu_1,\ldots,\mu_4$ (see Proposition \ref{prop:equiv}), and solve it iteratively via the gradient flow $\partial\mu_t  = -\nabla \c{SMW}^2(\mu_t, \nu_1,\ldots, \nu_P)$, following \textcite{bonneel} in the pairwise case.

\section{Multi-Task Learning with Sliced Multi-marginal Optimal Transport}

In the previous section, we proposed a multi-marginal metric between probability measures, which avoids exponential computational and statistical complexities and is thus practical for applications where a large number of samples $N$, number of measures $P$, or dimension $d$ is of interest. 
$\c{SMW}$ allows us to evaluate the closeness of probability measures $\mu_1,\ldots,\mu_P$, which makes it a good candidate regularizer in multi-task learning settings over probability spaces, by encouraging shared global structure across tasks through closeness in sliced multi-marginal geometry.
We now sketch potential areas of applications of $\c{SMW}$ in the context of multi-task learning on spaces of probability measures, and illustrate examples in density estimation and multi-dynamics reinforcement learning.

\subsection{Density Estimation with Shared Structure}
\label{sec:densityestimation}

Consider $P$ target measures $\mu_1, \ldots, \mu_P$, which we aim to approximate by parametric models $\nu_{_1}, \ldots,\nu_{P}$, such as for instance generative adversarial networks.
In applications, it is often the case that these measures are affected by issues related to \emph{distributional shift} \cite{distribshift}, which prevents us from obtaining accurate empirical samples of $\mu_1,\ldots,\mu_P$.
One way to counteract these issues is to introduce a shared structure between the measures, which can be enforced through $\c{SMW}$ regularization.

For example, consider empirical estimates $\h{\mu}_1,\ldots,\h{\mu}_P$ of $\mu_1,\ldots,\mu_P$, which are corrupted because no data is available in certain regions of each measure's support.
Here, reconstruction of $\mu_1,\ldots,\mu_P$ is only possible through the use of shared structure on the generative models $\nu_{1}, \ldots,\nu_{P}$, which we can enforce by using $\c{SMW}(\nu_{1}, \ldots,\nu_{P})$ as a regularizer.
This results in the optimization problem
\[
\argmin_{\nu_1,\ldots,\nu_P} \sum_{p=1}^P \underbracket[0.1ex]{\c{SW}^2(\mu_p, \nu_p)}_{\text{local loss}} +  \gamma \underbracket[0.1ex]{\c{SMW}^2(\nu_1,\ldots,\nu_P)}_{\text{global loss (shared)}},
\label{eq:multitask}
\]
where $\c{SW}^2(\mu_p, \nu_p)$ ensures that the respective generative models $(\nu_p)_{p=1}^P$ approximates targets $(\mu_p)_{p=1}^P$, and $\c{SMW}^2(\nu_{_1}, \ldots,\nu_{P})$ ensures shared structure is present in the loss.

\subsection{Multi-Dynamics Reinforcement Learning with Shared Structure}
\label{sec:multirl}

We now consider the problem of reinforcement learning in settings where the dynamics change. In order to speed up learning, we use  $\c{SMW}$ to share structure across different environments in this multi-dynamics reinforcement learning problem. 
Sharing knowledge is not only useful to bias (and thereby speed up) learning, but it is also useful in settings, where agents are ill informed, e.g., due to sparse reward signals. 
With a shared structure, these agents can learn from other agents. 
Here, the challenge is in effectively utilizing information from other agents in spite of differences in their respective environments. In the following, we focus on this setting.

Consider $P$ identical-task agents in finite-horizon Markov decision processes $(\c{S}, \c{A},\c{T}_p, r_p^{\text{env}})$, where $\c{S}$ is the state space and $\c{A}$ is the action space, both shared by all agents, $T_p(\v{x}^{(p)}_{t},\v{a}^{(p)}_t) = \v{x}^{(p)}_{t+1}$ is the  transition model of agent $p$, which varies across agents, and $r_p^{\text{env}}$ is the environment's reward function. 
Since different agents' tasks are identical, sharing structure can be beneficial. We consider the case, where some agents receive rewards $r_p^{\text{env}}=0$. These agents are \emph{uninformed} and can only learn  via a shared structure that allows to transfer knowledge from other agents.
Structure sharing is done by augmenting the agent-specific reward function  with a global multi-task reward term. In particular, define the augmented reward $R_p$ as
\[
\label{eq:rewards}
       R_p(\v{x}^{(p)}_t) = \underbracket[0.1ex]{r^{\text{env}}_p(\v{x}^{(p)}_t)}_{\substack{\text{agent specific}\\ \text{(local)}}} + \gamma \underbracket[0.1ex]{r^{\text{mul}}(\v{x}^{(p)}_t, \m{X})}_{\substack{\text{multi-task reward}\\\text{(shared/global)}}}, 
\]
where $\m{X} = \{\v{x}_t^{(p)}\}_{p,t=1}^{P,T}$ is the collection of all states of every agent at all time steps, $r^{\text{env}}_p(\v{x}^{(p)}_t)$ is the single-task reward of the $p^{th}$ environment and $r^{\text{mul}}(\v{x}^{(p)}_t, \m{X})$ is a (multi-task) reward signal.
The former provides task-specific information about the task to be solved by agent $p$, while the latter allows for agents to share structure through the history of their state trajectories. 
If $r^{\text{env}}_p = 0$ for a given agent, then this agent can only learn through the shared structure arising from the shared reward $r^{\text{mul}}$.  Finally, $\gamma$ is a regularizer that controls the influence of shared structure on the overall learning.

\begin{figure*}
\subfigure[Computational time (log-log scale, mean $\pm$ standard deviation over 5 runs) for computing the sliced multi-marginal distance in seconds against the number of samples for various $P$.]{
\includegraphics[width=0.3\hsize]{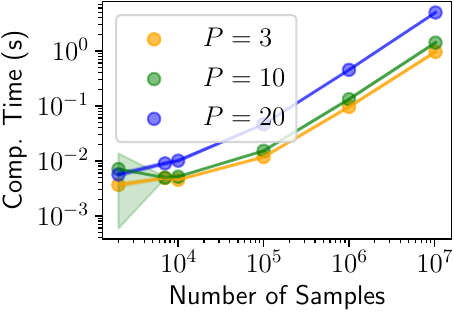}
\label{fig:compl_samples}
}%
\hfill
\subfigure[Computational time (log-log scale, mean $\pm$ standard deviation over 5 runs) for computing the sliced multi-marginal distance  in seconds against the number of measures, $d=10$ for various $N$.]{
\includegraphics[width=0.3\hsize]{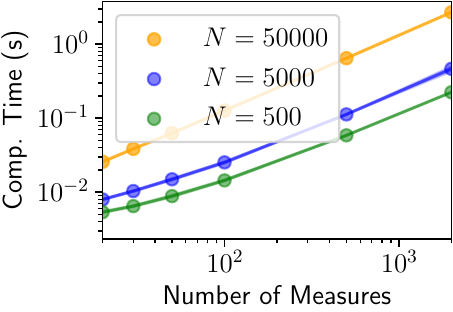}
 \label{fig:compl_mes}
}%
\hfill
\subfigure[Mean $\c{SMW}$ ($\pm$ standard deviations) sliced multi-marginal distance against the number of projections for $P=5$ measures with $N=250$ samples.]{
\includegraphics[width=0.3\hsize]{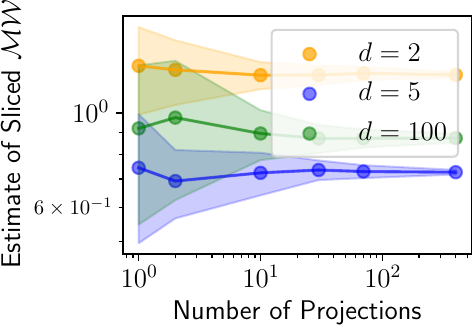}
\label{fig:accvar}
}
\caption{Properties of the sliced multi-marginal distance. \subref{fig:compl_samples} computational time as a function of the number of samples; \subref{fig:compl_mes} computational time as a function of the number of measures; \subref{fig:accvar} accuracy as a function of the number of projections}
\label{fig:analysis}
\end{figure*}

We now describe the shared reward $r^{\text{mul}}$. Denote $\mu_p = \frac{1}{T}\sum_{t=1}^T\delta_{\v{x}_t^{(p)}}$, which allows us to interpret the rollout of agent $p$ as a discrete probability measure supported on the state space. Then,
\[
 r^{\text{mul}}(\v{x}^{(p)}_t, \m{X})= - \frac{\beta_p}{K}\sum_{k=1}^K\Big|\langle \v{x}_t^{(p)}- \sum_{j=1}^P\beta_j \v{x}_{\eta_{p,j,k}(t)}^{(j)}, \v{\theta}_k\rangle \Big|^2,
\]
where $\eta_{p,j,k}$ returns the index of the atom in $\mu_j$ that is aligned with state $\v{x}_t^{(p)}$ after projecting on (Monte Carlo-sampled) $(\v{\theta}_k)_{k=1}^K$ and sorting all projected states. Intuitively, the reward signal attributed to the state $\v{x}^{(p)}_t$ of agent $p$ at time $t$ is computed by projecting all measures onto $K$ vectors, gathering all states that are aligned with $\v{x}^{(p)}_t$ for each projection $\v{\theta}_k$, and summing squared distances between them.

\begin{remark}
The barycentric cost structure with non-uniform  weights $\v{\beta}$ is particularly attractive in this setting, as it allows to give more weight to the communication arising from agents that perform well in their own environment. For instance, we can use Boltzmann weights 
\[
\beta_p\propto \exp\Big(\alpha\sum_{t=1}^{T}r_p^{\text{env}}(\v{x}^{(p)}_{t})\Big),
\]
where $\alpha$ is a temperature. It gives more weight in the  reward to agents performing best.
\end{remark}

We train all agents simultaneously by maximizing
\[\mathbb{E}_{\pi_1, \ldots, \pi_P}\Big[\sum_{p=1}^P\sum_{t=1}^T R_{p}(\v{x}^{(p)}_t)\Big] = \mathbb{E}_{\pi_1, \ldots, \pi_P} \Big[\sum_{p=1}^P\sum_{t=1}^T \underbracket[0.1ex]{r^{\text{env}}_{p}(\v{x}^{(p)}_t)  -\gamma\c{SMW}^2(\mu_1,\ldots, \mu_P)}_{=R_p(\v x_t^{(p)})}\Big] \label{eq:multirlobj}
\]
 with respect to the parameters of policies $\pi_p$, $p=1,\ldots,P$. 
Note that the extra term in the augmented reward regularizes the objective via the sliced multi-marginal Wasserstein distance. $\c{SMW}$ thus enforces closeness of  agents' trajectories which allows to share structure across agents.

\section{Experiments}
We now illustrate the behavior of sliced multi-marginal transport in simple multi-task learning setups.

\begin{figure}[t!]
    \centering
    \subfigure[$\gamma=0$ ]{
    \includegraphics[width=0.24\hsize]{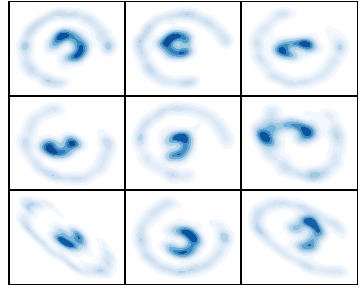}\label{fig:reg0}}
    \subfigure[$\gamma=0.3$]{\includegraphics[width=0.24\hsize]{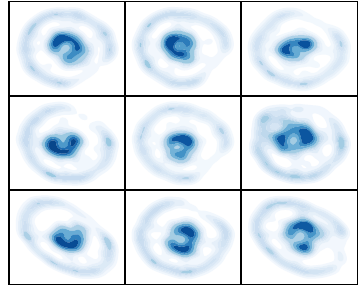}\label{fig:reg03}}
    \hspace{0.07cm}\subfigure[$\gamma=25$ ]{\includegraphics[width=0.24\hsize]{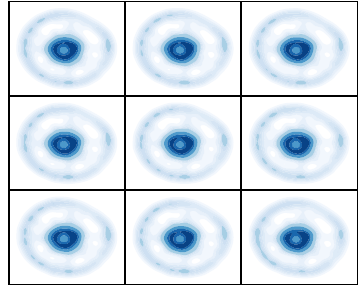}\label{fig:reg25}}
    \subfigure[Corrupted Targets]{\includegraphics[width=0.24\hsize]{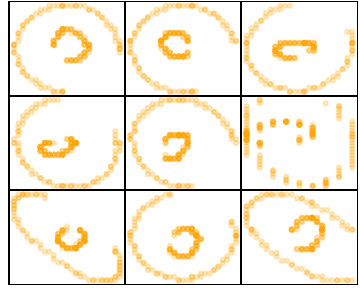}\label{fig:targets}}
    \caption{Multi-task density estimation experiment applied on corrupted nested ellipses (plotted in orange), using $\c{SW}^2$ as pairwise loss  and $\c{SMW}^2$ as regularizer. Learned models are plotted in blue. We use regularization coefficients $\gamma=0$ in \subref{fig:reg0}, $\gamma=0.3$ in \subref{fig:reg03}, $\gamma=25$ in \subref{fig:reg25}.}
    \label{fig:reg_all}
\end{figure}

\subsection{Scalability}


\textbf{Number of Samples ($N$).} We  study the impact of  the number of samples on the computational time to compute the sliced multi-marginal distance in \eqref{eq:smmwass}. In particular, we compute $\c{SMW}$ between $P=3, 10, 20$ measures in $\mathbb{R}^{10}$, $\mu_p\sim \c{N}(\v{m}_p,\eta^2 \m{I})$, where $p=1,\ldots, P$ for a fixed number of projections $K=10$. 
Figure \ref{fig:compl_samples} shows the   $\c{O}(N\log N)$ scaling of $\c{SMW}$. 
This enables computation of multi-marginal distances with over $10^7$ samples and a large number of measures.

\textbf{Number of Measures ($P$).} We now examine scaling with respect to the number of measures $P$. Figure \ref{fig:compl_mes} shows the time required to compute $\c{SMW}$ against $N=500, 5000, 50000$ measures. We observe the expected linear scaling of $\c{SMW}$.

\textbf{Number of Projections ($K$).} Finally, we consider the impact of the number of projections on the estimation of $\c{SMW}$ for dimensions $d=2,5, 20$. We set $N=250$, and $P=5$. Monte Carlo estimation is used to estimate $\c{SMW}$. Figure \ref{fig:accvar} shows the expected variance shrinkage as the number of projection grows, while the estimated mean converges to $\c{SMW}$ with rate $\c{O}(\frac{1}{\sqrt{K}})$ and constant factors depending on dimension.

\subsection{Multi-Task Density Estimation}

We consider the multi-task density estimation setting of Section \ref{sec:densityestimation}. Each target measures consist of a nested ellipse with corrupted samples. In particular, parts of each individual ellipse have been removed from each measure's support. Using the multi-task learning setup allows for sharing knowledge of the structure of the target tasks across problems---namely, that all target measures have the overall shape of nested ellipses. 
Figures \ref{fig:reg0}--\ref{fig:reg25} show the models obtained by multi-task training with regularization coefficients $\gamma = 0,\ 0.3,\ 25$. When $\gamma=0$, measures are learned individually without any structure sharing.  $\nu_1,\ldots, \nu_P$ hence collapse to the corrupted measures $\mu_1, \ldots ,\mu_P$. When structure is introduced ($\gamma>0$) knowledge of the inherent nested ellipse structure is shared across tasks, which leads to solutions that have such structure (holes are filled), but that still preserve the task-specific orientations and ellipse width/height as long as the structure coefficient $\eta$ is not too large. The latter causes the learned measures to be too close to each other. These effects can be seen in Figure \ref{fig:reg25}. When this happens, all learned measures collapse to the barycenter.

\subsection{Multi-Dynamics Reinforcement Learning}

\begin{figure}[t!]
\centering
\includegraphics[width=0.38\textwidth]{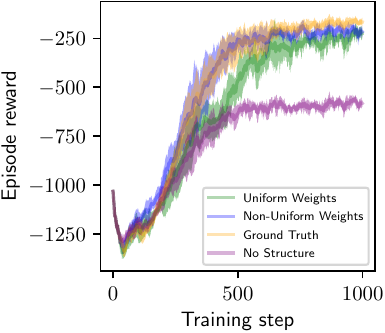}
\hspace{1cm}
\includegraphics[width=0.38\textwidth]{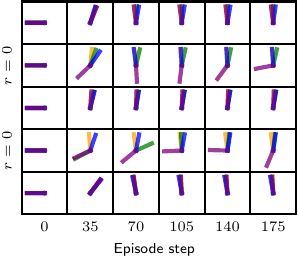}
\caption{Multi-task ($P=5$) RL experiment. Environments have different dynamics (different gravities), and  $2\slash 5$ agents have no environmental reward.  Without shared structure, these agents do not solve their respective tasks (orange). By contrast, with  shared structure, all agents learn accurate policies (green, blue), on par with agents trained without corrupted rewards (blue). Left: training curves (mean $\pm$ standard deviation averaged over $5$ runs), Right: states of agents for each task at the end of training (left to right refers to time $t$ from $0$ to $200$).}
\label{fig:pendulum results}
\end{figure}
We consider a multi-task RL application in the setting of Section \ref{sec:multirl}. In particular, we consider $P=5$ pendulum swing-up tasks with different dynamics (gravities $g\in\{8,9,10,11,12\}\,\text{m/s}^2$). States consist of angle and angular velocities, and actions of are torques. Environment rewards are dense as implemented in OpenAI Gym \cite{gym}, and following \textcite{Dadashi2020PrimalWI}, we transform the shared  reward $r^{\text{mul}}$ via $f(y) = e^{-5y}$. Two out of five agents do not receive any reward. All other agents share the same reward function. 
We consider agents trained with and without $\c{SMW}$-based regularization, and consider the uniform and non-uniform barycentric weights $\v{\beta}$; see Section \ref{sec:multirl} for more details. To facilitate learning, we lower-bound the weights of agents without reward. Policies are learned using $Q$-learning with function approximation on state observations.

Figure~\ref{fig:pendulum results} shows the results.
Training without regularization ($\gamma=0$, blue curve) does not allow the two agents without environment rewards ($r_p^{\text{env}}=0$) to solve their respective tasks. By contrast, with regularization, all agents (even those with no environment reward) solve their respective tasks (green, blue) as well as if all agents were receiving environmental rewards (orange). Agents with non-uniform regularization significantly outperform agents with uniform weights, showing that giving more weight in the regularizer to stronger agents is helpful. 
Overall, this demonstrates that knowledge transfer via the shared reward structure can be effective.
In particular, the regularization-based rewards encourage the state trajectories of all agents to be close under the sliced multi-marginal geometry. Hence, agents without environment rewards learn to \emph{follow} agents trained with environment rewards. This is possible because of similarity of environments and of agent goals, so that agent rollouts  share geometric structure.

\section{Conclusion}

In this work, we proposed a scalable multi-marginal optimal transport distance. Our main idea is to derive a closed-form formula for multi-marginal optimal transport in 1D in the general case and to extend it into a higher-dimensional metric via slicing. We show it is well-behaved topologically, and in particular that it is a generalized metric. We also show it is well-behaved statistically with dimension-free sample complexity (modulo a caveat arising from projection complexity). We derive a range of other results illustrating the simple and intuitive geometric structure of sliced multi-marginal transport. Finally, we propose areas of applications of sliced multi-marginal transport in the context of multi-task learning on probability spaces, and concrete instantiations in density estimation, and reinforcement learning. We hope these contributions enable practitioners in reinforcement learning, generative modeling and other areas to share structure across tasks in a geometrically-motivated way. Our work relies on the assumption that tasks live on the same space, and share structure. Future work extends our approach to allow for multi-task learning on incomparable spaces, enabling structure sharing in more general set-ups, for instance via Gromov--Wasserstein-like techniques.

\section*{Acknowledgments}
SC was supported by the Engineering and Physical Sciences Research Council (grant number EP/S021566/1).

\printbibliography

\newpage
\appendix

\section{Proofs}

\subsection{Closed-form Formulas for Multimarginal Optimal Transport}
\label{sec:closedformsproof}

For a measure $\mu \in \c{M}(\R)$, define its CDF $C_\mu:\R\-> [0,1]$ as
\[
    C_\mu(x) = \int_{-\infty}^x d\mu(y) \ \forall x.
\]
Also, define its pseudo-inverse $C_\mu^{-1}:[0,1]\-> \R \cup  \{-\infty\}$ as
\[
    C^{-1}_\mu(r) = \min_x \{x \in \R\cup \{-\infty\}:C_\mu(x)\geq r\}.
\]
This function is a generalization of the quantile function.

\paragraph{1D Multi-Marginal}

\begin{customprop}{\ref{prop:closed_form}}
\label{prop:eq_wass_kant}
If $\mu_1,\ldots, \mu_P \in \c{M}(\mathbb{R})$ and $\c{U}(0,1)$ is the uniform measure, then
\[
        \mathcal{MW}^2(\mu_1,\ldots,\mu_P)&=\int_{0}^1\sum_{p=1}^{P}\beta_p\Big|C_{\mu_p}^{-1}(x)-\sum_{j=1}^P\beta_jC_{\mu_j}^{-1}(x)\Big|^2\d x,
\]
and the optimal coupling solving \eqref{eq:multi_marginal_w} is of the form
\[
        \pi^\star = (C_{\mu_1}^{-1},\ldots, C_{\mu_P}^{-1})_{\#}\ \c{U}(0,1).
\]\end{customprop}

\begin{proof}
Our aim is to provide a closed form formula for
\[
    \mathcal{MW}^2(\mu_1,\ldots,\mu_P)&=\min_{\pi \in \Pi(\mu_1,\ldots, \mu_P)} \int_{(\R^d)^P} \sum_{p=1}^P \beta_p |  x_p - \sum_j\beta_jx_j| ^2 d\pi(x_1,\ldots,x_P),
\]
where $\Pi(\mu_1,\ldots, \mu_P)$ is the set of probability measures in $\c{M}((\R^d)^P)$ with marginals $\mu_1, \ldots, \mu_P$.

First, notice
\[
     &\quad \int_{(\R^d)^P} \sum_{p=1}^P \beta_p \Vert  x_p - \sum_j\beta_j x_j\Vert ^2 d\pi(x_1,\ldots,x_P)\\
     &=\sum_{p=1}^P \beta_p \int_{\R^d} |x_p|^2d\mu_p -2 \sum_{p,j=1}^P \beta_p\beta_j \int_{(\R^d)^2} x_px_j\d\pi_{pj}(x_p,x_j),
\]
where $\pi_{pj}$ corresponds to marginalizing $\pi$ onto all components but $p,j$. This can be formalized by defining the map $L_{pj}(x_1,\ldots,x_P) = (x_p,x_j)$ and $\pi_{pj} = L_{pj\#}\pi$.

Now define $\pi^\star= (C_{\mu_1}^{-1},\ldots, C_{\mu_P}^{-1})_{\#}\mathcal{U}(0,1)$
\paragraph{Claim:} $\pi^\star$ is optimal

First observe $L_{pj\#}\pi^\star = (C_{\mu_p}^{-1}, C_{\mu_j}^{-1})_{\#}\mathcal{U}(0,1)$ by marginalization. Note this is the optimal coupling between pairs $\mu_p , \mu_j$, see \cite{compopt} (this can easily be obtained by observing that plugging in $(C_{\mu_p}^{-1},C_{\mu_j}^{-1})_{\#}\mathcal{U}(0,1)$ into the Wasserstein objective achieves the minimum -- it is also a valid coupling, thus it has to be the optimal coupling.)

Now, note that 
\[\argmax_{\gamma \in \Pi(\mu_p,\mu_j)}\int_{(\R^d)^2} x_px_j\d\gamma = \argmin_{\gamma \in \Pi(\mu_p,\mu_j)}\int_{(\R^d)^2}|x_p-x_j|^2\d\gamma,
\]
and also that for any multimarginal coupling $\pi\in \Pi(\mu_1,\ldots, \mu_P)$, $\pi_{pj}$ is a pairwise coupling in $\Pi(\mu_p,\mu_j)$ by the transfer lemma.

We can hence deduce that $\forall \pi\in \Pi(\mu_1,\ldots, \mu_P)$
\[
\int_{(\R^d)^2} x_px_jd\pi_{pj}\leq \int_{(\R^d)^2} x_px_jd\pi^\star_{pj} \quad \forall p,j =1,\ldots, P,
\]
because both $\pi_{pj}$ and $\pi^\star_{pj}$ are couplings of $\mu_p, \mu_j$ and $\pi^\star_{pj}$ is optimal.

Therefore, it holds that
\[
     &\quad \int_{(\R^d)^P} \sum_{p=1}^P \beta_p \Vert  x_p - \sum_j\beta_j x_j\Vert ^2 \d\pi^\star(x_1,\ldots,x_P)
     \\
     &=\sum_{p=1}^P \beta_p \int_{\R^d} |x_p|^2d\mu_p -2 \sum_{p,j=1}^P \beta_p\beta_j \int_{(\R^d)^2} x_px_j\d\pi^\star_{pj}(x_p,x_j)
     \\
     &\leq \sum_{p=1}^P \beta_p \int_{\R^d} |x_p|^2d\mu_p -2 \sum_{p,j=1}^P \beta_p\beta_j \int_{(\R^d)^2} x_px_j\d\pi_{pj}(x_p,x_j)
     \\
     &= \int_{(\R^d)^P} \sum_{p=1}^P \beta_p \Vert  x_p - \sum_j\beta_j x_j\Vert ^2 \d\pi(x_1,\ldots,x_P),
\]
which proves the claim that $\pi^\star$ is the optimal multi-marginal coupling. We now compute the distance by plugging in the optimal coupling:
\[
    \mathcal{MW}^2(\mu_1,\ldots,\mu_P)
    &=\int_{(\R^d)^P} \sum_{p=1}^P \beta_p |  x_p - \sum_j\beta_jx_j| ^2 \d\pi^\star(x_1,\ldots,x_P)
    \\
     &=\int_{(\R^d)^P} \sum_{p=1}^P \beta_p |  x_p - \sum_j\beta_jx_j| ^2 \d(C_{\mu_1}^{-1},\ldots,C_{\mu_p}^{-1})_{\#}\mathcal{U}(0,1)
     \\
    &= \int_{0}^1 \sum_{p=1}^P \beta_p |  C_{\mu_p}^{-1}(x) - \sum_j\beta_jC_{\mu_j}^{-1}(x)| ^2 \d x.
\]
\end{proof}

\subsection{Generalized Metric Properties}
\label{sec:metricproofs}

\begin{definition}
\label{def:mw_prop}
Assume $\mu_p \in \c{M}(\R^d)$, where  $p~=~1,\ldots, P$, and let $D:\c{M}(\R^d)\times \ldots \times \c{M}(\R^d) \-> \mathbb{R} $ be a multi-marginal Wasserstein metric with barycentric weights $\v{\beta}$. Then, $D$ is a generalized metric if the following properties hold:
\begin{list}{\labelitemi}{\leftmargin=0.25em}
   \addtolength{\itemsep}{-.215\baselineskip}
    \item[$1$.] $\begin{aligned}[t] &D(\mu_1,\ldots,\mu_P) &\geq 0 \end{aligned}$ \label{pro:1}
    \item[$2$.] $\begin{aligned}[t] &D(\mu_1,\ldots,\mu_P) & = 0  \Leftrightarrow \mu_1 =\ldots = \mu_P \end{aligned}$\label{pro:2}
    \item[$3$.] $\begin{aligned}[t] &D(\mu_1,\ldots,\mu_P) & =D_\sigma(\mu_{\sigma(1)},\ldots,\mu_{\sigma(P)}), \ \forall \sigma \in \mathbb{S}_P \end{aligned}$ where $D_\sigma$ denotes that the barycentric weights $\v{\beta}$ are permuted by $\sigma$ and $\mathbb{S}_P$ is the group of permutations of order $P$. \label{pro:3}
    \item[$4$.] $\begin{aligned}[t] &\forall \mu \in \c{M}(\R^d):D(\mu_1,\ldots,\mu_P)                \leq  \sum_{p=1}^P D(\mu_1,\ldots,\mu_{p-1},\mu,\mu_{p+1},\ldots  \mu_{P})& \end{aligned} $\label{pro:4}
\end{list}
\end{definition}

\begin{proposition}
\label{prop:mw_prop_cont}
$\c{MW}$ is a generalized metric on the restriction $\c{M}(\mathbb{R})$.
\end{proposition}

\begin{proof}
Property (1), i.e., positivity is clear because
\[
\mathcal{MW}^2(\mu_1,\ldots,\mu_P)&=\int_{0}^1\sum_{p=1}^{P}\beta_p\Big|C_{\mu_p}^{-1}(x)-\sum_{j=1}^P\beta_jC_{\mu_j}^{-1}(x)\Big|^2\d x\geq0
\]

Next, we prove property (2).

We begin by proving the forward implication ($\Rightarrow$).
\begin{align}
&\quad \c{MW}(\mu_1,\ldots,\mu_P) = 0 
\\
&\Rightarrow \Big(\int_{0}^1\sum_{p=1}^{P}\beta_p\Big|C_{\mu_p}^{-1}(x)-\sum_{j=1}^P\beta_jC_{\mu_j}^{-1}(x)\Big|^2\d x\Big)^{\frac{1}{2}}  = 0
\\ 
&\Rightarrow \int_{0}^1\sum_{p=1}^{P}\beta_p\Big|C_{\mu_p}^{-1}(x)-\sum_{j=1}^P\beta_jC_{\mu_j}^{-1}(x)\Big|^2\d x  = 0
\\ 
&\Rightarrow C_{\mu_p}^{-1}(x)-\sum_{j=1}^P\beta_jC_{\mu_j}^{-1}(x)  = 0 \quad \forall p = 1,\ldots, P,\  \forall x \in [0,1]
\end{align}
Now assume for contradiction that $ \exists m, n, x: \quad C_{\mu_m}^{-1}(x) \neq C_{\mu_n}^{-1}(x)$, then:
\begin{align}
  &\quad C_{\mu_m}^{-1}(x) = \sum_{j=1}^P\beta_jC_{\mu_j}^{-1}(x),\quad \quad  \quad  C_{\mu_n}^{-1}(x) = \sum_{j=1}^P\beta_jC_{\mu_j}^{-1}(x)
  \\
  &\Leftrightarrow C_{\mu_m}^{-1}(x) - C_{\mu_n}^{-1}(x) = \sum_{j=1}^P\beta_jC_{\mu_j}^{-1}(x) - \sum_{j=1}^P\beta_jC_{\mu_j}^{-1}(x) = 0
\end{align}
which is a contradiction, therefore $C_{\mu_m}^{-1}(x) = C_{\mu_n}^{-1}(x) \quad \forall m,n,x$, thus $\mu_1 = \ldots = \mu_P$ 

We continue by proving the backward implication ($\Leftarrow$).

If $\mu_1=\ldots = \mu_P$, then $C_{\mu_p}^{-1}(x)= C_{\mu_{p'}}^{-1}(x) \quad \forall x, \ \forall p,p'=1,\ldots,P$. 

Therefore, $C_{\mu_p}^{-1}(x)-\sum_{j=1}^P\beta_jC_{\mu_j}^{-1}(x)  = 0 \quad \forall p=1,\ldots,P, \ \forall x \in [0,1]$.
Thus, 
\begin{align}
    &\quad \c{MW}(\mu_1,\ldots,\mu_P)=\Big(\int_0^1\sum_{p=1}^P\beta_p|C_{\mu_p}^{-1}(x)-\sum_{j=1}^P\beta_jC_{\mu_j}^{-1}(x)|^2dx\Big)^{\frac{1}{2}}  =0.
\end{align}
We continue with permutation invariance (3),
\begin{align}
    \c{MW}(\mu_1,\ldots,\mu_P) &= \Big(\int_{0}^1\sum_{p=1}^{P}\beta_p\Big|C_{\mu_p}^{-1}(x)-\sum_{j=1}^P\beta_jC_{\mu_j}^{-1}(x)\Big|^2\d x\Big)^{\frac{1}{2}} \\
    &=\Big(\int_{0}^1\sum_{p=1}^{P}\beta_p\Big|C_{\mu_p}^{-1}(x)-\sum_{j=1}^P\beta_{\sigma(j)}C_{\mu_{\sigma(j)}}^{-1}(x)\Big|^2\d x\Big)^{\frac{1}{2}}
    \\
    &=\Big(\int_{0}^1\sum_{p=1}^{P}\beta_{\sigma(p)}\Big|C_{\mu_{\sigma(p)}}^{-1}(x)-\sum_{j=1}^P\beta_jC_{\mu_{\sigma(j)}}^{-1}(x)\Big|^2\d x\Big)^{\frac{1}{2}}\\
    &=\c{MW}_\sigma(\mu_{\sigma(1)},\ldots,\mu_{\sigma(P)})
\end{align}
Equalities holds because sums are invariant under any permutation $\sigma$.

We finally prove the generalized triangle inequality (4). Note the slight abuse of notation that $p+1$ component does not exist when $p=P$.

We begin by proving the case $P\geq 3$.
Firstly, we rewrite the multi-marginal functional in the following way:
\begin{align}
    \c{MW}^2(\mu_1,\ldots,\mu_P)& = \sum_{p=1}^{P}\beta_p\int_{0}^1\Big|C_{\mu_p}^{-1}(x)-\sum_{j=1}^P\beta_jC_{\mu_j}^{-1}(x)\Big|^2\d x
    \\
    & = \frac{1}{2}\sum_{p,p'=1}^{P}\beta_p\beta_{p'}\int_{0}^1\Big|C_{\mu_p}^{-1}(x)-C_{\mu_{p'}}^{-1}(x)\Big|^2\d x
    \\
    & = \frac{1}{2}\sum_{p,p'=1}^{P}\beta_p\beta_{p'}\int_{0}^1f_{p,p'}^2(x)\d x
\end{align}
where $f_{p,p'}(x) = \Big|C_{\mu_p}^{-1}(x)-C_{\mu_{p'}}^{-1}(x)\Big|$. The results holds because
\begin{align} \sum_{m,n=1}^P \beta_m \beta_n |C_{\mu_m}^{-1}(x)-C_{\mu_n}^{-1}(x)|^2&= \sum_{m=1}^P \beta_m \Big|C_{\mu_m}^{-1}(x)-\sum_{n=1}^P \beta_n C_{\mu_j}^{-1}(x)\Big|^2,
\end{align}
which holds because 
\begin{align}
    &\sum_{m=1}^P \beta_m \Big|x_m-\sum_{n=1}^P \beta_n x_n\Big|^2
    \\
    &= \sum_{m=1}^P \beta_m \Big[|x_m|^2 + |\sum_{n=1}^P \beta_n x_n|^2 -2\sum_{n=1}^P \beta_n x_m x_n\Big]
    \\
    &= \sum_{m=1}^P \beta_m |x_m|^2 + \sum_{m,n=1}\beta_m\beta_n x_m x_n -2\sum_{m,n=1}^P \beta_m\beta_n x_m x_n
    \\
    &= \sum_{m=1}^P \beta_m |x_m|^2 - \sum_{m,n=1}\beta_m\beta_n x_m x_n
    \\
    &= \sum_{m,n=1}^P \beta_m \beta_n |x_m|^2 - \sum_{m,n=1}\beta_m\beta_n x_m x_n
    \\
    &= \sum_{m,n=1}^P \beta_m \beta_n  ( \frac{1}{2}|x_m|^2 + \frac{1}{2}|x_n|^2 -  x_m x_n )
    \\
    &= \frac{1}{2}\sum_{m,n=1}^P \beta_m \beta_n  |x_m -x_n|^2 .
\end{align}
Therefore, we have
\begin{align}
    \sum_{p=1}^P \c{MW}^2(\mu_1,\ldots,\mu_{p-1},\mu, \mu_{p+1},\ldots,\mu_P) &= \frac{1}{2}\sum_{p=1}^P \sum_{m,n\neq p}^{P} \beta_m\beta_{n}\int_{0}^1f_{n,m}^2(x)dx + C,
\end{align}
where $C>0$.

We now show that $\int_{0}^1\sum_{p=1}^P \sum_{m,n\neq p}^{P} \beta_m\beta_{n}f_{n,m}^2(x)dx \geq \sum_{p,p'=1}^{P}\beta_p\beta_{p'}\int_{0}^1f_{p,p'}^2(x)\d x$. 
This can be  observed by noting that all $\int_0^1f_{p,p'}^2(x)\d x$ terms on the RHS appear on the LHS. Indeed, for any $m',n'$, $\int_0^1f_{m',n'}^2(x)\d x$ appears in the $p \neq m',n'$ summation, which always holds for some $p$ as $P\geq 3$.

Therefore, we have shown that
\begin{align}
    \c{MW}^2(\mu_1,\ldots,\mu_P) \leq \sum_{p=1}^P \c{MW}^2(\mu_1,\ldots,\mu_{p-1}, \mu, \mu_{p+1},\ldots,\mu_{P}) 
\end{align}
Also,
\begin{align}
    \quad \c{MW}^2(\mu_1,\ldots,\mu_P) & \leq \sum_{p=1}^P \c{MW}^2(\mu_1,\ldots,\mu_{p-1},\mu, \mu_{p+1},\ldots,\mu_{P})   
    \\
    \Rightarrow  \c{MW}(\mu_1,\ldots,\mu_P) & \leq \sqrt{\sum_{p=1}^P \c{MW}^2(\mu_1,\ldots,\mu_{p-1},\mu, \mu_{p+1},\ldots,\mu_{P}) }
    \\
    &\leq   \sum_{p=1}^P \sqrt{ \c{MW}^2(\mu_1,\ldots,\mu_{p-1},\mu, \mu_{p+1},\ldots,\mu_{P}) }
    \\
    &=\sum_{p=1}^P \c{MW}(\mu_1,\ldots,\mu_{p-1},\mu, \mu_{p+1},\ldots,\mu_{P}) 
\end{align}
which proves the result. The case $P=2$  has been proved via different approaches (e.g. \cite{compopt}).

\end{proof}

\begin{customprop}{\ref{prop:metricprop}}
$\c{SMW}$ is a generalized metric on the restriction $\c{M}(\mathbb{R}^d)$.
\end{customprop}

\begin{proof}

Property (1) holds by definition due to positivity of $\c{MW}$ on $\R$ and the definition of the sliced multi-marginal distance.

Property (2) is more delicate. We begin with the forward direction ($\Rightarrow$).

We extend the proof of \textcite{sl_prop} to the multi-marginal case. Define $\Theta$ as the uniform distribution on $S_{d-1}$.   Define `for ($\Theta$-almost-every) $\v{\theta}$' as $\forall \Theta \text{-a-e-} \v{\theta}$. Firstly, the following holds:
\begin{align} &\quad \c{SMW}(\mu_1,\ldots,\mu_P) = 0\\
&\Rightarrow \Big( \frac{1}{\text{Vol}(S_{d-1})}\int_{S_{d-1}} \c{MW}^2(M_{\v{\theta}\#}\mu_1,\ldots,M_{\v{\theta}\#}\mu_P)d\Theta(\v{\theta})    \Big)^{\frac{1}{2}} = 0\\
&\Rightarrow  \c{MW}(M_{\v{\theta}\#}\mu_1,\ldots,M_{\v{\theta}\#}\mu_P) \ \forall \Theta \text{-a-e-} \v{\theta} \\
&\Rightarrow M_{\v{\theta}\#}\mu_1 = \ldots = M_{\v{\theta}\#}\mu_P \ \forall \Theta \text{-a-e-} \v{\theta}
\end{align}

Next, we define the Fourier transform of any measure $\mu$ on $\c{M}(\mathbb{R}^s), s\geq 1$ at any $\v{w} \in \mathbb{R}^s$:
\begin{align}
    \c{F}[\mu](\v{w}) = \int_{\mathbb{R}^s}e^{-i\langle \v{w}, \v{x} \rangle}\d\mu(\v{x}).
\end{align}

Therefore, using properties of push-forwards, the following holds:
\begin{align}
     \c{F}[M_{\v{\theta}\#}\mu](t) = \int_{\mathbb{R}}e^{-itu}dM_{\v{\theta}\#}\mu(u) = \int_{\mathbb{R}^s}e^{-it\langle \v{\theta}, \v{x} \rangle}d\mu(\v{x}) = \c{F}[\mu](t\v{\theta}).
\end{align}

As  $\forall \Theta \text{-a-e-} \v{\theta}$, $M_{\v{\theta}\#}\mu_1 = \ldots = M_{\v{\theta}\#}\mu_P$, then $\c{F}[M_{\v{\theta}\#}\mu_1]=\ldots=\c{F}[M_{\v{\theta}\#}\mu_P]$, which implies that $\c{F}[\mu_1]=\ldots=\c{F}[\mu_P]$. By injectivity of the Fourier transform, we conclude that $\mu_1=\ldots=\mu_P$.

We continue with the backward direction ($\Leftarrow$).

We assume $\mu_1=\ldots = \mu_P$, which implies the following:
\begin{align}
    &\quad \mu_1=\ldots = \mu_P\\
    &\Rightarrow M_{\theta\#}\mu_1 = \ldots = M_{\theta\#}\mu_P\ \ \forall \Theta \text{-a-e-} \v{\theta}\\
    &\Rightarrow \c{MW}^2(M_{\v{\theta}\#}\mu_1,\ldots,M_{\v{\theta}\#}\mu_P) = 0 \ \ \forall \Theta \text{-a-e-} \v{\theta}\\
    &\Rightarrow \c{SMW}(\mu_1,\ldots,\mu_P) = \Big( \frac{1}{\text{Vol}(S_{d-1})}\int_{S_{d-1}} \c{MW}^2(M_{\v{\theta}\#}\mu_1,\ldots,M_{\v{\theta}\#}\mu_P)\d\Theta(\v{\theta})     \Big)^{\frac{1}{2}} = 0 .
\end{align}

We now prove Property (3)
\begin{align}
    \c{SMW}(\mu_1,\ldots,\mu_P) &= \Big( \frac{1}{\text{Vol}(S_{d-1})}\int_{S_{d-1}} \c{MW}^2(M_{\v{\theta}\#}\mu_1,\ldots,M_{\v{\theta}\#}\mu_P)\d\Theta(\v{\theta})     \Big)^{\frac{1}{2}}
    \\
    &=\Big( \frac{1}{\text{Vol}(S_{d-1})}\int_{S_{d-1}} \c{MW}_\sigma^2(M_{\v{\theta}\#}\mu_{\sigma(1)},\ldots,M_{\v{\theta}\#}\mu_{\sigma(P)})d\Theta(\v{\theta})     \Big)^{\frac{1}{2}}
    \\
    &= \c{SMW}_{\sigma}(\mu_{\sigma(1)},\ldots,\mu_{\sigma(P)})
\end{align}

We finally end by proving Property (4), the generalized triangle inequality.

Earlier, we showed that
 \begin{align}
     \c{MW}^2(\mu_1,\ldots,\mu_P) \leq \sum_{p=1}^P \c{MW}^2(\mu_1,\ldots,\mu_{p-1},\mu, \mu_{p+1},\ldots,\mu_{P}) .
 \end{align}
 This implies that
 \begin{align}
 &\quad \c{SMW}^2(\mu_1,\ldots,\mu_P)\\
     &= \frac{1}{\text{Vol}(S_{d-1})}\int_{S_{d-1}} \c{MW}^2(M_{\v{\theta}\#}\mu_1,\ldots,M_{\v{\theta}\#}\mu_P)\d\Theta(\v{\theta}) 
     \\
     &\leq \sum_{p=1}^P \frac{1}{\text{Vol}(S_{d-1})}\int_{S_{d-1}} \c{MW}^2(M_{\v{\theta}\#}\mu_1,\ldots,M_{\v{\theta}\#}\mu_{p-1},M_{\v{\theta}\#}\mu, M_{\v{\theta}\#}\mu_{p+1},\ldots,M_{\v{\theta}\#}\mu_{P})\d\Theta(\v{\theta}) 
     \\
     &=\sum_{p=1}^P \c{SMW}^2(\mu_1,\ldots, \mu_{p-1},\mu, \mu_{p+1},\ldots, \mu_{P}).
 \end{align}
Therefore, we conclude that
\begin{align}
   &\quad  \c{SMW}^2(\mu_1,\ldots,\mu_P) \leq \sum_{p=1}^P \c{SMW}^2(\mu_1,\ldots, \mu_{p-1},\mu, \mu_{p+1},\ldots, \mu_{P})\\ 
   & \Rightarrow \c{SMW}(\mu_1,\ldots,\mu_P) \leq \sum_{p=1}^P \c{SMW}(\mu_1,\ldots, \mu_{p-1},\mu, \mu_{p+1},\ldots, \mu_{P})
\end{align}
directly in the same way as in the proof of Proposition the generalized triangle inequality for $\c{MW}$.
\end{proof}

\subsection{Mathematical Properties}
\begin{customprop}{\ref{prop:meansliced}}
\label{prop:equal_apdx}
\[
\c{SMW}^2(\mu_1, \ldots, \mu_P)  = \frac{1}{2} \sum_{i,j=1}^P\beta_i\beta_j \c{SW}^2(\mu_i,\mu_j)
\] 
\end{customprop}

\begin{proof}

\[
    \c{SMW}^2(\mu_1,\ldots,\mu_P) &= \frac{1}{\text{Vol}(S_{d-1})}\int_{S_{d-1}} \int_{\R^d}\frac{1}{2}\sum_{i,j=1}^P\beta_i\beta_j|x_i-x_j|^2\d\pi^{\star \v{\theta}}(x_1,\ldots,x_P) \d\Theta(\v{\theta}) 
    \\
    &= \frac{1}{2\text{Vol}(S_{d-1})}\sum_{i,j=1}^P\beta_i\beta_j\int_{S_{d-1}}\int_{\R \times \R}|x_i-x_j|^2\d\pi_{ij}^{\star \v{\theta}}(x_i,x_j) \d\Theta(\v{\theta}) 
    \\
    &= \frac{1}{2}\sum_{i,j=1}^P\beta_i\beta_j\frac{1}{\text{Vol}(S_{d-1})}\int_{S_{d-1}}\c{W}^2(M_{\v{\theta}\#}\mu_i, M_{\v{\theta}\#}\mu_j) \d\Theta(\v{\theta}) ,
\]
where $\pi^{\star \v{\theta}}$ is the optimal coupling between $M_{\v{\theta}\#}\mu_1,\ldots, M_{\v{\theta}\#}\mu_P$ and $M_{\v{\theta}}(\v{x}) = \innerprod{\v{x}}{\v{\theta}}$. Similarly to proofs of closed-form formulas for multi-marginal Kantorovich transport, we know that $\pi_{ij}^{\star \v{\theta}}$ is the optimal coupling between $M_{\v{\theta}\#}\mu_i, M_{\v{\theta}\#}\mu_j$. As a result, it holds that
\[
    \c{SMW}^2(\mu_1,\ldots,\mu_P)&=\frac{1}{2}\sum_{i,j=1}^P\beta_i\beta_j\c{SW}^2(\mu_i,\mu_j).
\]

\end{proof}

\begin{corollary}
\[
\c{SMW}^2(\mu_1, \ldots, \mu_P)  \leq \frac{1}{2} \sum_{i,j=1}^P\beta_i\beta_j \c{W}^2(\mu_i,\mu_j)
\] 
\end{corollary}

\begin{proof}
By Proposition \ref{prop:equal_apdx}, it holds that 
\[
    \c{SMW}^2(\mu_1,\ldots,\mu_P)&=\frac{1}{2}\sum_{i,j=1}^P\beta_i\beta_j\c{SW}^2(\mu_i,\mu_j).
\]
Also, \textcite{bonnottee} shows that
\[
\c{SW}^2(\mu,\nu) \leq \c{W}^2(\mu,\nu)  \ \  \forall \mu, \nu.
\]
The result follows directly.
\end{proof}

\begin{corollary}
\[0 \leq a^{2(d+1)}  \sum_{i,j =1}^P \beta_i \beta_j \c{W}^{4(d+1)}(\mu_i, \mu_j) \leq b^{2(d+1)}  \c{SMW}^2(\mu_1,..., \mu_P).\]
\label{cororeverse}
\end{corollary}

\begin{proof}
\textcite{bonnottee} has shown that it holds for some positive constants $a,b$ that 

\[0 \leq a  \c{W}^2(\mu_i, \mu_j) \leq b  \c{SW}^{1/(d+1)}(\mu_i, \mu_j), \]

and that $x^{d+1}$ is an increasing function for all positive $x$. Therefore, raising both sides to the power of $2(d+1)$, we obtain that
\[0 \leq a^{2(d+1)}  \c{W}^{4(d+1)}(\mu_i, \mu_j) \leq b^{2(d+1)}  \c{SW}^2(\mu_i, \mu_j).\]

Now summing across $i,j$, and weighting with the barycentric cost’s weights, we obtain 
\[0 \leq a^{2(d+1)}  \sum_{i,j =1}^P \beta_i \beta_j \c{W}^{4(d+1)}(\mu_i, \mu_j) \leq b^{2(d+1)}  \c{SMW}^2(\mu_1,..., \mu_P).\]

It therefore follows that as $\c{SMW}^2(\mu_1,..,\mu_P) \to 0$, we also have $\c{W}^{4(d+1)}(\mu_i, \mu_j) \to 0$ for each pair of measures, and hence by positivity of $\c{W}$ that $\sum_{i,j = 1}^P \beta_i \beta_j \c{W}^2(\mu_i,\mu_j) \to 0$. 

\end{proof}

\subsection{Sample/Projection Complexity}
\label{sec:complexitiesproofs}
We now study $E [\c{SMW}^2(\mu_1,\ldots,\mu_P) - \c{SMW}^2(\h{\mu}_1,\ldots,\h{\mu}_P)]^2$ where $\h{\mu}_p$'s refers to empirical measures with $n$ samples.  Then the following result holds:
\begin{customprop}{\ref{prop:samplecomplexity}}
\label{prop:samples}
If $\mu_1,\ldots,\mu_P \in \c{M}(\R^d)$, and assuming $\c{W}^2$ has sample complexity $\rho(N)$ on $\R$, then
\[ &\quad E [\c{SMW}^2(\mu_1,\ldots,\mu_P) - \c{SMW}^2(\h{\mu}_1,\ldots,\h{\mu}_P)]^2 
\leq\frac{1}{2}\rho(N).\]
\end{customprop}
This result shows the sample complexity is dimension free.

\begin{proof}
We conclude from Proposition \ref{prop:meansliced}
\[
    \c{SMW}^2(\mu_1,\ldots,\mu_P) -\c{SMW}^2(\h{\mu}_1,\ldots,\h{\mu}_P) &=\frac{1}{2}\sum_{i,j=1}^P\beta_i\beta_j\Big(\c{SW}^2(\mu_i,\mu_j)-\c{SW}^2(\h{\mu}_i,\h{\mu}_j)\Big).
\]
If $\c{W}^2$ on $\R$ has sample complexity $\rho(N)$, then $\c{SW}^2$ on $\R^d$ also has sample complexity $\rho(N)$, i.e., its sample complexity is dimension free. The proof relies on an application of Jensen's inequality and is a special case of \textcite{topstatprop}.

\[
E \Big|\c{SW}^2(\mu,\nu) - \c{SW}^2(\hat{\mu}_n,\hat{\nu}_n)\Big| &= E \Big|\int_{S_{d-1}} \{\c{W}^2(\theta^{*}_{\#}\mu,\theta^{*}_{\#}\nu)-\c{W}^2(\theta^{*}_{\#}\hat{\mu}_n,\theta^{*}_{\#}\hat{\nu}_n)\} \,d\Theta(\theta)\Big| \\
&\leq E \left\{\int_{S_{d-1}} \Big|\c{W}^2(\theta^{*}_{\#}\mu,\theta^{*}_{\#}\nu)-\c{W}^2(\theta^{*}_{\#}\hat{\mu}_n,\theta^{*}_{\#}\hat{\nu}_n)\Big| \,d\Theta(\theta)\right\} \\
&\leq \int_{S_{d-1}} E \Big|\c{W}^2(\theta^{*}_{\#}\mu,\theta^{*}_{\#}\nu)-\c{W}^2(\theta^{*}_{\#}\hat{\mu}_n,\theta^{*}_{\#}\hat{\nu}_n)\Big| \,d\Theta(\theta) \\
&\leq \int_{S_{d-1}} \rho(N)\,d\Theta(\theta) = \rho(N)
\]

Hence,
\[
    &\quad E\Big|\c{SMW}^2(\mu_1,\ldots,\mu_P) -\c{SMW}^2(\h{\mu}_1,\ldots,\h{\mu}_P)\Big|
    \\
    &=E\Big|\frac{1}{2}\sum_{i,j=1}^P\beta_i\beta_j\Big(\c{SW}^2(\mu_i,\mu_j)-\c{SW}^2(\h{\mu}_i,\h{\mu}_j)\Big)\Big|
    \\
    &\leq \frac{1}{2}\sum_{i,j=1}^P\beta_i\beta_jE\Big|\c{SW}^2(\mu_i,\mu_j)-\c{SW}^2(\h{\mu}_i,\h{\mu}_j)\Big|
    \\
    &\leq \frac{1}{2}\sum_{i,j=1}^P\beta_i\beta_j \rho(N) = \frac{1}{2}\rho(N) .
\]
\end{proof}

Here we also derive similar results to theirs about projection complexity. 
\begin{customprop}{\ref{prop:projectioncomplexity}}
Let $\mu_1,\ldots,\mu_P \in \c{M}(\R^d)$, and define $\widebar{\c{SMW}}$ the approximation obtained by uniformly picking $L$ projections on $S_{d-1}$, then
\[ 
\E\left[ \widebar{\c{SMW}}^2(\mu_1,\ldots,\mu_P) - \c{SMW}^2(\mu_1,\ldots,\mu_P)\right]^2
\leq 
L^{-1/2}\Var_{\v{\theta}}\Big[\c{MW}^2\big(\mu_1^{\v{\theta}},\ldots,\mu_P^{\v{\theta}})\Big],
\]
where  $\v{\theta}$ follows the uniform distribution on $S_{d-1}$ and $\mu_p^{\v{\theta}} = M_{\#}^{\v{\theta}}(\mu_p)$.
\end{customprop}

\begin{proof}
We bound the error arising from the Monte Carlo approximation of $\c{SMW}$, similarly to \textcite{topstatprop} in the pairwise case. In particular, define $\delta =\int_{S_{d-1}} \c{MW}^2(M_{\theta\#}\mu_1,\ldots,M_{\theta\#}\mu_P)\d\Theta(\theta)$. Then we have that
\[
&\quad E_{\v{\theta}\sim \sigma} |\widebar{\c{SMW}}^2(\mu_1,\ldots,\mu_P) - \c{SMW}^2(\mu_1,\ldots,\mu_P)|
\\
&\leq \Big\{E_{\v{\theta}\sim \sigma}
|\widebar{\c{SMW}}^2(\mu_1,\ldots,\mu_P) - \c{SMW}^2(\mu_1,\ldots,\mu_P)|^2
\Big\}^{\frac{1}{2}}
\\
&\leq L^{-1\slash2}\int_{S_{d-1}} \Big\{\c{MW}^2(M_{\theta\#}\mu_1,\ldots,M_{\theta\#}\mu_P)-\delta \Big\}^2\d\Theta(\theta)\\
&=L^{-1/2}\Var_{\v{\theta}}\Big[\c{MW}^2\big(\mu_1^{\v{\theta}},\ldots,\mu_P^{\v{\theta}})\Big],\]
which holds due to the same Monte-Carlo concentration inequality as in \textcite{topstatprop} (Proof of Theorem 6).
\end{proof}

\subsection{Equivalence to Sliced Barycenters and Weak Convergence}
\label{sec:equivalenceproofs}

\begin{customprop}{\ref{prop:equiv}}
Let $\mu_1,\ldots,\mu_P\in \c{M}(\mathbb{R}^d)$, $\sum_{p=1}^P \beta_p=1$. Furthermore, let $\hat{\beta}_p$ be augmented multi-marginal weights, so that for $m \in [0,1]$ it holds that $\hat{\beta}_p = m \beta_p$ for $p=1,\ldots,P$, $\sum_{p=1}^{P+1}\hat{\beta}_p = 1$, and $\c{D}=\c{SW}^2$. Then
\[
    \argmin_{\mu \in \c{M}(\mathbb{R}^d)} \c{SMW}^2(\mu_1,\ldots,\mu_P, \mu) = \argmin_{\mu \in \c{M}(\mathbb{R}^d)}\c{F}(\mu),
\]
where $\v{\beta}$ is the weight vector of $\c{F}$ and $\v{\hat{\beta}}$ is the weight vector of $\c{SMW}$.
\end{customprop}
\begin{proof} 
\[
   &\quad \argmin_{\mu \in \c{M}(\mathbb{R}^d)}\c{SMW}^2(\mu_1,\ldots,\mu_P, \mu)\\
   &=\argmin_{\mu \in \c{M}(\mathbb{R}^d)} \sum_{p=1}^P\hat{\beta}_p\hat{\beta}_{P+1}\c{SW}^2(\mu, \mu_p)\\
   &=\argmin_{\mu \in \c{M}(\mathbb{R}^d)} \sum_{p=1}^P\beta_p\c{SW}^2(\mu_p,\mu)
   \\
   &=\argmin_{\mu \in \c{M}(\mathbb{R}^d)} \c{F}(\mu).
  \]
\end{proof}

\subsection{Differentiability}

\begin{customprop}{\ref{prop:differentiability}}
Let $\mu_1, \ldots, \mu_P \in \c{M}(\R^d)$ be discrete measures with $N$ atoms, which we gather into matrices $\{\m{X}^{(p)}\}_{p=1}^P$, and similarly define $\mu_{\m{X}}$ with atoms $\m{X}$. Assume $\m{X}$ has distinct points. Then $\c{SMW}^2$ is smooth with gradient
\[
\nabla_{\m{X}} \c{SMW}^2(\mu_1,\ldots,\mu_P, \mu_{\m{X}}) =\beta_{P+1}\sum_{p=1}^{P} \beta_{p}\int_{S_{d-1}}\m{X}_{\v{\theta}}-\big(\m{X}_{\v{\theta}}^{(p)} \circ \sigma_{\m{X}_{\v{\theta}}} \circ \sigma^{-1}_{\m{X}^{(p)}_{\v{\theta}}} \big) \d\Theta(\v{\theta}),
\]
where $\sigma_{\m{X}}$ is the permutation that sorts atoms of $\m{X}$.
\end{customprop}

\begin{proof}
Define $\sigma_{\m{Y}}$ be the permutation of $\{1,\ldots,N\}$ that sorts atoms of $\m{Y}$. Also, define $\m{X}_{\v{\theta}} \in \R^N $, such that $(\m{X}_{\v{\theta}})_i = \innerprod{\v{x}_i}{\v{\theta}}$. Then
\[
\c{SMW}^2(\mu_1,\ldots,\mu_P, \mu_{\m{X}}) &= \sum_{p=1}^P\beta_{P+1} \beta_{p}\c{SW}^2(\mu_{\m{X}},\mu_p) + C(\mu_1,\ldots,\mu_P).
\]
Hence, 
\[
\nabla_{\m{X}} \c{SMW}^2(\mu_1,\ldots,\mu_P, \mu_{\m{X}}) &=\nabla_{\m{X}}\sum_{p=1}^P\beta_{P+1} \beta_{p}\c{SW}^2(\mu_{\m{X}},\mu_p)
\\
&= \sum_{p=1}^{P}\beta_{P+1} \beta_{p}\int_{S_{d-1}}\m{X}_{\v{\theta}}-\m{X}_{\v{\theta}}^{(p)} \circ\big( \sigma_{\m{X}_{\v{\theta}}} \circ \sigma^{-1}_{\m{X}^{(p)}_{\v{\theta}}}\big)  \d\v{\theta}.
\]
The last equality is due to \textcite{bonneel}.
\end{proof}
\clearpage
 \section{Additional Experimental Details}
 
 We now provide further experimental details. All experiments ran on CPU, besides the benchmarking experiments, which ran on a single P100 GPU.
 
 \textbf{Ellipses - Multi-Task Density Estimation} 
 
 We set the batch size to $150$, and parametrize each measure $\nu_p$ as a discrete measure with $150$ atoms which we optimize over via stochastic gradient descent. We set the number of projections to $20$. 
 
 \textbf{Multi-Task Reinforcement Learning} 
 
 The horizon is set to $T=200$. The learning rate is set to $2.5\times 10^{-4}$, and the batch size to optimize the $Q$-function to $32$. The Q-network is a $2$-layer MLP with $\tanh$ activation. We use $f(x) = e^{-5x}$ to rescale the reward function following \textcite{Dadashi2020PrimalWI}, we set the number of projections to $K=50$ and $\gamma=1$. Also, we set $\alpha = \frac{1}{30}$. Our implementation extends the repository \url{https://github.com/xtma/simple-pytorch-rl} to the multi-task setting, and leverages OpenAI gym environments \cite{gym}.
 
 \textbf{Gradient Flow experiment}
 
 We follow the setup of \textcite{bonneel}. In particular, we discretize the flow to numerically estimate it via gradient descent $\m{X}^{(l+1)} = \m{X}^{(l+1)} - \nabla \c{SMW}^2(\mu_1,\ldots,\mu_P, \mu_{ \m{X}^{(l)}})$, and plot the location of particles for $l=0,\ldots,T$ where T is the number of steps (200), which approximates the gradient flow. We estimate $\c{SMW}$ with $30$ projections. Each measure (including the initial measure $\mu_0$ consist in samples from isotropic Gaussians, and the initial measure.

\end{document}